\PassOptionsToPackage{table}{xcolor}
\PassOptionsToPackage{hyphens}{url}
\documentclass[10pt,a4paper]{article}
\usepackage{zju-ant-preprint}
\usepackage[authoryear,round]{natbib}
\usepackage{subcaption,multirow,array,colortbl}

\renewcommand{\bibfont}{\small}
\renewcommand{\PaperTitle}{CAVC: Convergence-Adaptive Audio-Visual Cache for  Modality-Asymmetric Diffusion Generation}
\renewcommand{\PaperShortTitle}{CAVC: Convergence-Adaptive Audio-Visual Cache}
\renewcommand{\PaperAuthors}{Lifan Jiang\textsuperscript{1,2},\quad Yibo Zhao\textsuperscript{1,2},\quad Senlin cheng\textsuperscript{2} \\[1pt]
  Bolei Xu\textsuperscript{2},\quad XIAOBO LI\textsuperscript{2},\quad Can Ye\textsuperscript{2} \\[1pt]
  Dehai Zhao\textsuperscript{1},\quad Deng Cai\textsuperscript{1},\quad Boxi Wu\textsuperscript{1,\textdagger}}
\renewcommand{\PaperAffiliations}{\textsuperscript{1}State Key Lab of CAD\&CG, Zhejiang University\qquad\textsuperscript{2}Ant Group}
\renewcommand{\PaperContact}{lifanjiang@zju.edu.cn}
\renewcommand{\PaperAuthorNote}{\textsuperscript{\textdagger}Corresponding author: Boxi Wu.}
\renewcommand{\PaperKeywords}{Audio-visual generation; Diffusion models; Adaptive caching; Inference acceleration}

\renewcommand{\PaperDate}{October 2026}
\hypersetup{pdfauthor={Lifan Jiang, Yibo Zhao, Senlin cheng, Bolei Xu, XIAOBO LI, Can Ye, Dehai Zhao, Deng Cai, Boxi Wu}}
\renewcommand{\PaperAbstract}{%
Diffusion-based audio-visual generation methods typically perform joint denoising of audio and video latents throughout the entire sampling trajectory, repeatedly executing modality-specific denoising branches and cross-modal attention at every timestep. However, we observe that such uniform joint computation is not always necessary. The key is a modality-asymmetric recoverability: audio-side drift introduced by mid-stage caching can be substantially reduced by only a few terminal joint denoising steps, whereas video-side degradation caused by caching leads to persistent loss of motion and temporal details. This asymmetry suggests that inference acceleration should selectively cache recoverable audio-side computation while preserving full video refinement.
Motivated by this observation, we propose \textbf{Convergence-Adaptive Audio-Visual Cache (CAVC)}, a training-free modality-asymmetric caching framework. CAVC adopts a sandwich-structured computation allocation: after detecting that audio denoiser outputs enter a cache-tolerant regime, it caches and reuses the audio denoiser output to bypass audio denoising and cross-modal attention in the middle stage, while the video branch continues independent spatio-temporal refinement; a few joint steps are then restored at the end to reduce accumulated cross-modal drift and improve synchronization. Experiments across multiple public benchmarks demonstrate that CAVC achieves 1.10--1.35$\times$ wall-clock acceleration, improves several visual dynamics and temporal statistics, and maintains competitive audio fidelity and audio-visual alignment.
}

\begin{document}
\MakeJointTitle

\begin{figure*}[t]
\centering
\includegraphics[width=0.95\textwidth]{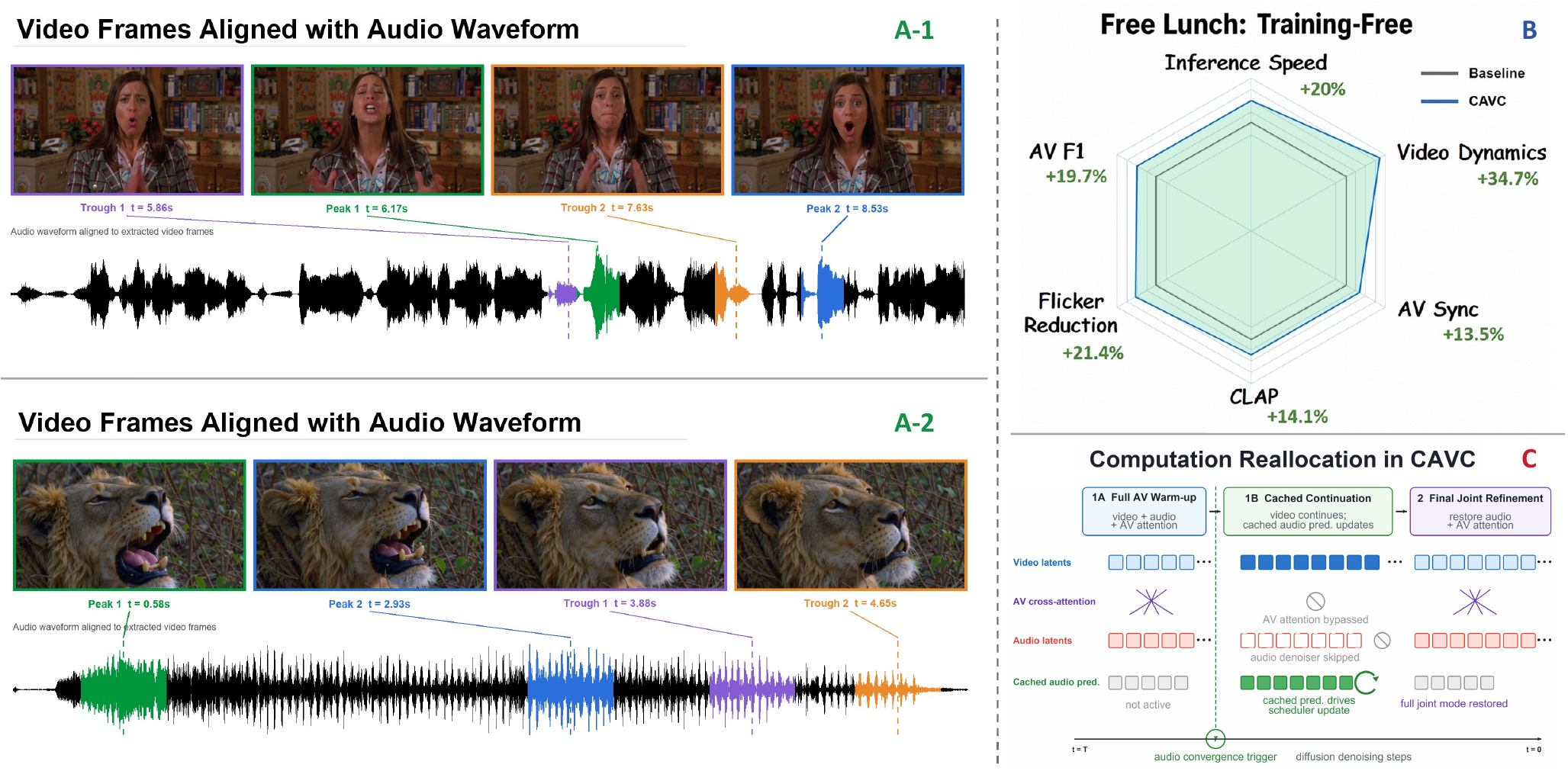}
    
    \caption{\textbf{A:} CAVC generates high-fidelity video frames synchronized with complex audio waveforms. \textbf{B:} On General-80, training-free CAVC achieves a favorable quality-efficiency trade-off across generation quality, cross-modal alignment, and measured wall-clock inference speed. \textbf{C:} CAVC exploits modality-asymmetric recoverability: mid-stage audio-side drift induced by caching can be substantially reduced by a short terminal joint refinement, whereas video-side degradation is difficult to recover. CAVC preserves full video refinement while bypassing recoverable audio-side computation in the middle stage.}
    \label{fig:teaser}
    \vspace{1.3em}
\end{figure*}

\section{Introduction}

Text-to-audio-visual generation aims to synthesize semantically consistent video and temporally synchronized audio from text, becoming a pivotal task in multimodal generative modeling \cite{singer2022make, ho2022video, blattmann2023align, huang2023make, chen2025foley, wang2026unifiedav}. While unimodal diffusion models excel independently \cite{guo2023animatediff, chen2023videocrafter1, wang2023audioldm2}, their temporal and semantic alignment requires joint denoising via audio-video cross-modal attention \cite{ruan2023mm, tang2024codi, tang2024codi2, xing24seeing, luo2023difffoley}. Despite its effectiveness, this paradigm introduces prohibitive inference costs by repeatedly executing modality-specific branches and dense cross-attention at every sampling step, creating a major bottleneck for large-scale deployment.

A straightforward remedy is to adopt existing diffusion acceleration techniques, including advanced ODE/SDE solvers \cite{song2020denoising, lu2022dpm, zhao2023unipc}, knowledge distillation for few-step generation \cite{salimans2022progressive, luo2023lcm, meng2023distillation, yin2024one}, and training-free feature caching across adjacent timesteps \cite{ma2024deepcache, zhao2023param, wimbauer2024cache}. However, directly applying these strategies to audio-visual generation is not always satisfactory. Reducing the entire sampling trajectory compresses refinement of both modalities, often harming video details and temporal dynamics. Meanwhile, modality-agnostic caching \cite{ma2024deepcache} treats audio and video computation uniformly, ignoring their potentially different roles throughout denoising. These limitations raise a fundamental question: \emph{is full audio-video joint computation necessary at every denoising step?}

We answer this question through the lens of a modality-asymmetric recoverability in joint audio-visual denoising. Importantly, we do not assume that audio predictions numerically converge faster than video predictions. Instead, CAVC exploits a recoverability asymmetry: audio-side drift induced by mid-stage caching can be substantially reduced by a few terminal joint steps, whereas video-side degradation leads to persistent loss of motion and temporal details that brief terminal recoupling cannot recover. This asymmetry is supported by our branch-freezing ablations, where caching or freezing the video branch yields high raw speedups but severely degrades video dynamics, while audio-only caching preserves video refinement and benefits from joint tuning.

Motivated by this observation, we propose \textbf{Convergence-Adaptive Audio-Visual Cache} (\textbf{CAVC}), a training-free modality-asymmetric caching framework with a sandwich-structured computation allocation. Once the audio denoiser output enters a cache-tolerant regime, CAVC caches and reuses the audio-side network output to bypass the audio denoising branch in subsequent middle steps; the audio latent is still updated by the original scheduler using the cached output, so its sampling trajectory continues without repeated audio forward passes. Simultaneously, CAVC bypasses audio-video cross-modal attention, allowing the video branch to proceed with independent spatio-temporal refinement conditioned on its visual state and textual guidance. Finally, CAVC restores full joint denoising for a few terminal steps. Empirically, this short recoupling reduces cross-modal drift and improves audio-visual alignment before decoding.

We also provide a conditional perturbation analysis of cached-audio and video-only residuals. Under local smoothness assumptions, the bounds expose how these residuals and cache duration affect deviation from the full sampler; recoverability is evaluated empirically. On AVGen-Bench, VABench T2AV, and General-80, CAVC achieves 1.10--1.35$\times$ speedups while improving visual dynamics and maintaining competitive audio fidelity and audio-visual alignment.
Our main contributions are summarized as follows:
\begin{itemize}
    
    \item \textbf{Modality-asymmetric recoverability.} We identify that audio-side drift introduced by mid-stage caching can be substantially reduced by a few terminal joint denoising steps, whereas video-side degradation is difficult to recover, motivating asymmetric computation allocation.

    \item \textbf{Sandwich-structured asymmetric caching.} We propose CAVC, a training-free framework that bypasses recoverable audio-side computation and cross-modal attention in the middle stage while preserving full video refinement, then restores minimal joint denoising at the end for synchronization refinement.

    \item \textbf{Favorable quality-efficiency trade-off.} Across three benchmarks, CAVC achieves 1.10--1.35$\times$ acceleration, improves several motion and temporal metrics, and maintains competitive audio fidelity and audio-visual alignment, supported by ablations and perturbation analysis.
\end{itemize}

\begin{figure*}[t]
    \centering

    \includegraphics[width=0.94\textwidth]{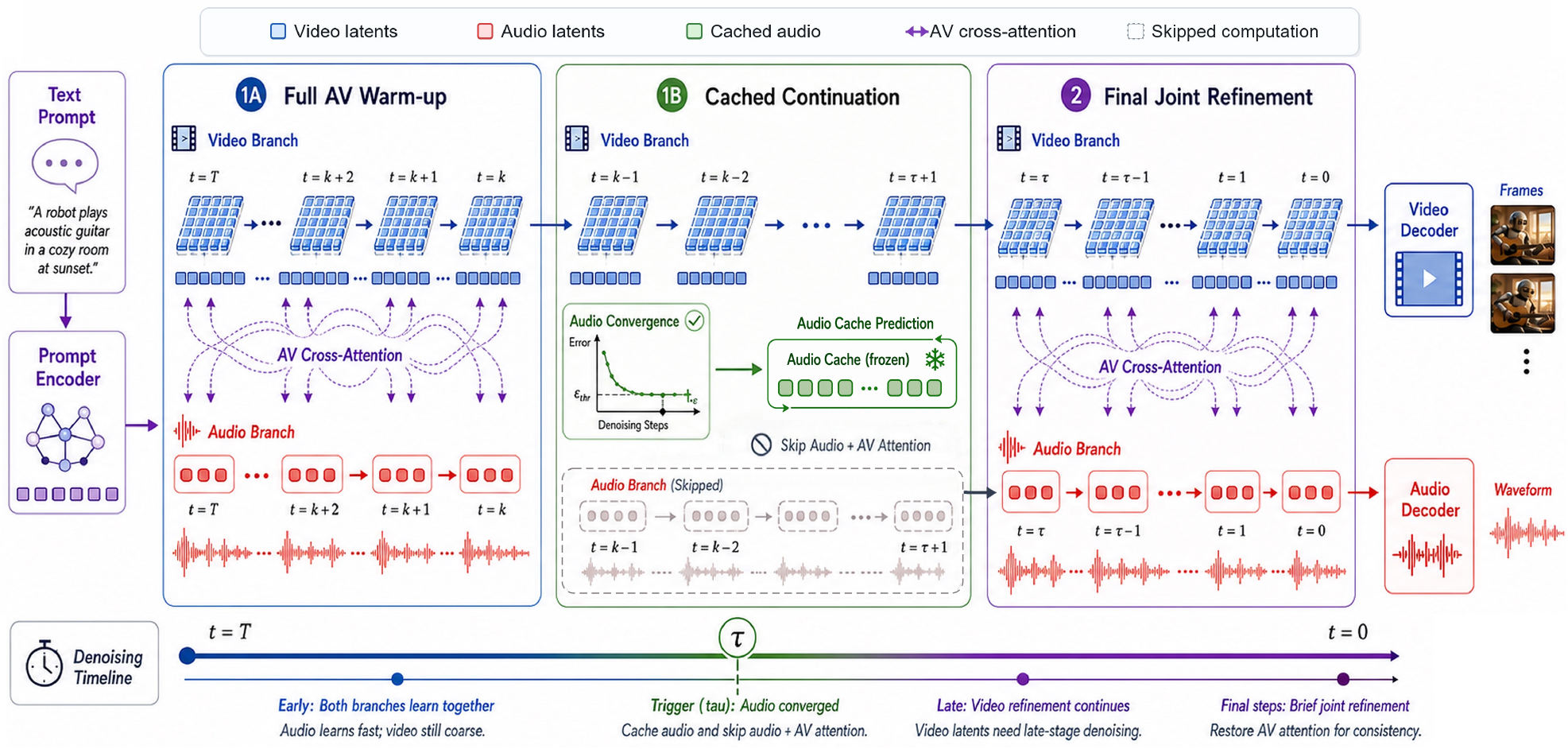} 
    
    \caption{Overview of the CAVC framework. The denoising trajectory ($t=T \to 0$) is dynamically partitioned based on modality convergence. \textbf{(1A) Full AV Warm-up:} During early high-noise stages, audio/video branches and AV cross-attention fully execute to establish semantic binding. \textbf{(Trigger $\tau$):} An online monitor tracks audio prediction changes; once stabilized, the audio cache activates. \textbf{(1B) Cached Continuation:} Cached audio predictions are reused. Bypassing the expensive audio branch and AV attention frees the video branch for independent spatio-temporal refinement without being over-constrained by static audio. \textbf{(2) Final Joint Refinement:} Full joint computation is terminally restored for a few steps to reduce accumulated cross-modal drift and refine audio-visual synchronization before decoding.}
    
    \label{fig:pipeline}
\end{figure*}

\section{Related Work}

\subsection{Audio-Visual Diffusion Generation}
Multimodal generative modeling has advanced from independent unimodal video \cite{singer2022make, guo2023animatediff} and audio synthesis \cite{liu2023audioldm, wang2023audioldm2} to audio-visual joint generation. Representative frameworks such as MM-Diffusion \cite{ruan2023mm} and CoDi \cite{tang2024codi} achieve this by passing audio and video latents through dual denoising branches fused via dense cross-modal attention. Recent advancements \cite{xing2024seeing, klear2026unified, lin2026crossmodal} employ latent aligners to enhance fine-grained synchronization. However, this tightly coupled paradigm mandates repeated execution of modality-specific branches and expensive cross-modal interactions \cite{zhang2025dynamic} at every denoising step, causing prohibitive computational overhead. Our work addresses this bottleneck by dynamically reallocating computation during inference, bypassing the need for architectural redesigns.

\subsection{Acceleration of Diffusion Models}
Accelerating diffusion inference predominantly reduces sampling steps via advanced solvers \cite{song2020denoising, lu2022dpm, zhao2023unipc, liu2025rectified} or distillation \cite{luo2023lcm, yin2024one, meng2023distillation, zheng2025step}. Complementary training-free approaches reduce per-step costs via token merging \cite{bolya2023tokenmergingvitfaster, gao2024token} or structured pruning \cite{fang2023structuralpruningdiffusionmodels}. Applying these trajectory-shortening techniques to audio-visual generation uniformly truncates critical spatio-temporal refinement, inevitably degrading visual dynamics. In contrast, CAVC preserves necessary sampling steps for the video modality, selectively approximating recoverable audio-side computation while maintaining the video refinement trajectory.

\subsection{Modality-Agnostic Feature Caching}

Orthogonal to step reduction, training-free mechanisms exploit temporal redundancy across adjacent timesteps to avoid redundant evaluations. Methods like DeepCache \cite{ma2024deepcache} cache high-level features, while recent extensions explore evolutionary scheduling \cite{ecad2025evolutionary}, adaptive video caching \cite{fastercache2024, team2024chameleon}, and temporal redundancy minimization \cite{wu2025temporalcache}. Despite their success, existing caching strategies remain inherently modality-agnostic, treating audio and video uniformly with synchronous caching operations. 
This uniform treatment overlooks the inherent recoverability asymmetry between modalities. CAVC differentiates itself by redistributing computation according to this asymmetry, applying convergence-adaptive caching exclusively to the recoverable audio branch.

\section{Method}

We propose CAVC, a training-free acceleration framework for asymmetric dual-stream audio-visual diffusion transformers, selectively approximating recoverable audio-side computation while preserving full video refinement.

\subsection{Asymmetric Problem Formulation}

\label{sec:formulation}

We consider a standard text-conditioned audio-visual diffusion model. Given text condition $c$, let $v_t$ and $a_t$ denote the video and audio latents at denoising step $t\in\{T,T-1,\ldots,1\}$ with noise level $\sigma_t$. Larger $t$ corresponds to higher noise, and denoising proceeds from $t=T$ to the final state at $t=0$. For empirical trigger locations, we use the zero-based forward execution index $k=T-t$, which increases as $t$ decreases. The full joint denoising model simultaneously predicts denoised outputs for both modalities:
\begin{equation}
    \hat{y}_v^t = f_v(v_t, a_t, c, \sigma_t), \qquad
    \hat{y}_a^t = f_a(v_t, a_t, c, \sigma_t),
\end{equation}
where $f_v$ and $f_a$ denote the video and audio branches, mutually coupled through audio-to-video (A2V) and video-to-audio (V2A) cross-modal attention. The scheduler subsequently updates both latents:
\begin{equation}
    v_{t-1} = S_t^v\!\left(v_t,\, \hat{y}_v^t\right), \qquad
    a_{t-1} = S_t^a\!\left(a_t,\, \hat{y}_a^t\right).
\end{equation}

Executing both branches at every step supports semantic binding but introduces severe computational redundancy. Our core observation is a recoverability asymmetry between modalities. When we refer to audio stabilization, we do not imply that audio predictions numerically converge faster than video predictions. Rather, we refer to the onset of a cache-tolerant regime—detectable via an online criterion—where reusing audio denoiser outputs introduces drift that remains empirically reducible by terminal joint refinement. By recoverability, we mean the empirical property that such audio-side drift can be substantially reduced by a few terminal joint denoising steps. Video dynamics do not share this property: interrupting video refinement causes persistent motion and temporal degradation that brief terminal recoupling cannot restore. This asymmetry provides the motivation for CAVC: inference acceleration must maintain full video-branch activity, approximate only recoverable audio-side computation, and briefly restore joint denoising at the end.

\subsection{Convergence-Adaptive Audio-Visual Cache}
\label{sec:cavc}

Driven by the recoverability asymmetry, CAVC adopts a sandwich-structured computation allocation with three phases: a full joint warm-up to establish cross-modal binding, an asymmetric cached continuation phase that bypasses recoverable audio-side computation while video continues full refinement, and a brief final joint refinement to reduce cross-modal drift.

\noindent\textbf{Stage 1 (Phase A): Full Audio-Visual Warm-up.}
For $T\geq t\geq\tau$, CAVC follows the full sampler above, establishing scene layout, audio categories, and cross-modal binding. Premature decoupling can introduce drift that terminal refinement does not reliably remove.

\begin{figure}[t]
    \centering
    \includegraphics[width=0.9\linewidth]{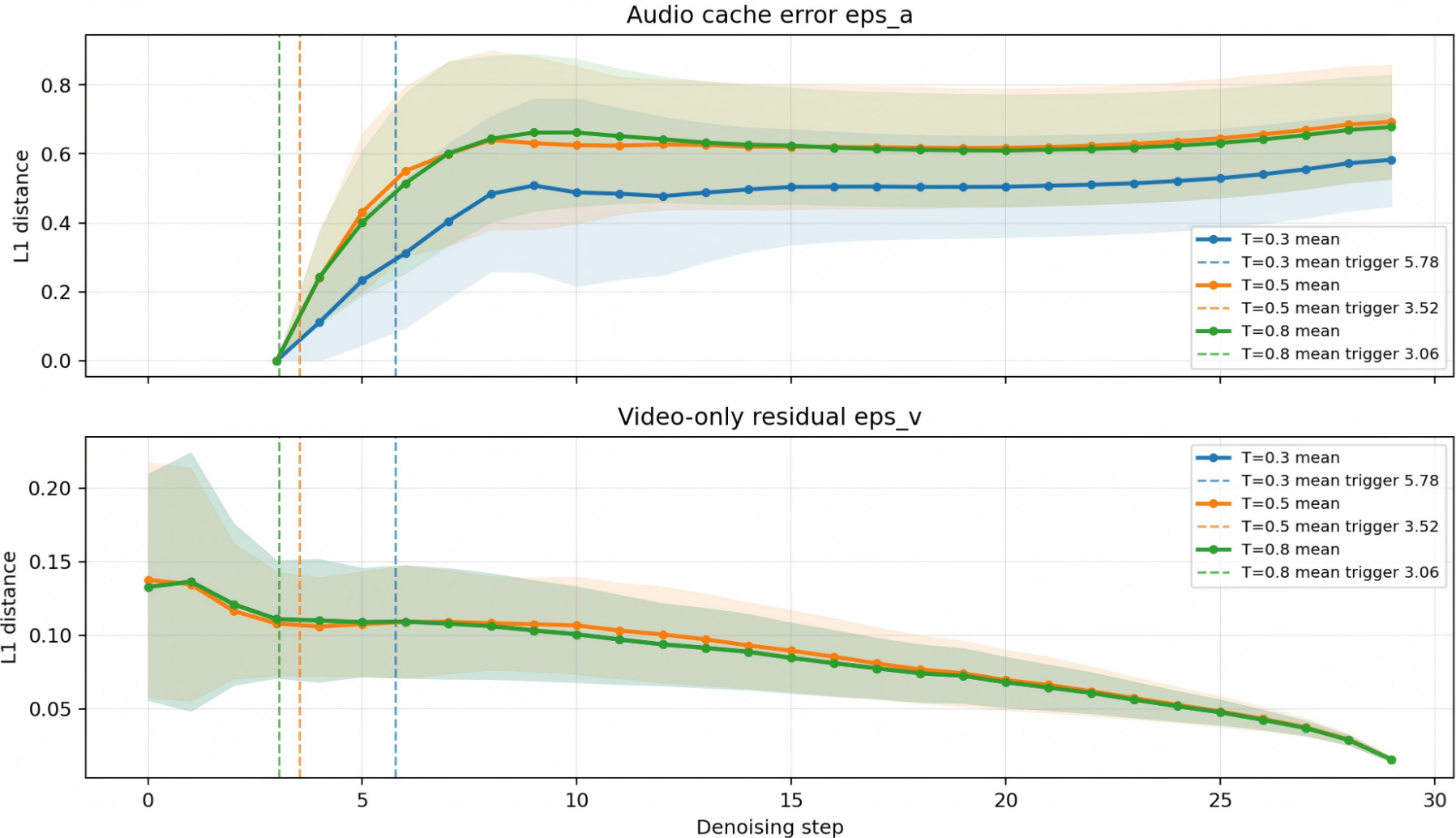}
    
    \caption{Empirical approximation residuals over the forward execution index $k$. \textbf{Top:} The instantaneous audio cache residual grows after cache triggering (dashed lines) but stays within a stable empirical range; a tighter threshold $T$ ($\theta$ in text; e.g., 0.3) delays the trigger and lowers the observed residual ceiling. \textbf{Bottom:} The video-only residual stabilizes and decays, suggesting that the dominant cross-modal dependencies are established during Phase A warm-up.}
        
    \label{fig:convergence_errors}
\end{figure}

\noindent\textbf{Stage 1 (Trigger): Online Audio Convergence Detection.}
Because audio cacheability varies with prompt complexity, fixed triggers are suboptimal. CAVC thus employs adaptive online monitoring; for $t<T$, it computes the mean absolute change from the previously executed, higher-noise prediction at $t+1$:
\begin{equation}
    \Delta_a^t = \mathrm{mean}\!\left(\left|\hat{y}_a^t - \hat{y}_a^{t+1}\right|\right).
\end{equation}
Caching triggers at timestep $\tau$ when $\Delta_a^t < \theta$ persists for $W$ execution steps. This threshold is an empirical proxy for a cache-tolerant regime (Figure~\ref{fig:convergence_errors}, Table~\ref{tab:ablation_master}d), after which CAVC stores $\hat{y}_a^*=\hat{y}_a^\tau$.

\begin{figure*}[t]
    \centering
    \includegraphics[width=0.94\textwidth]{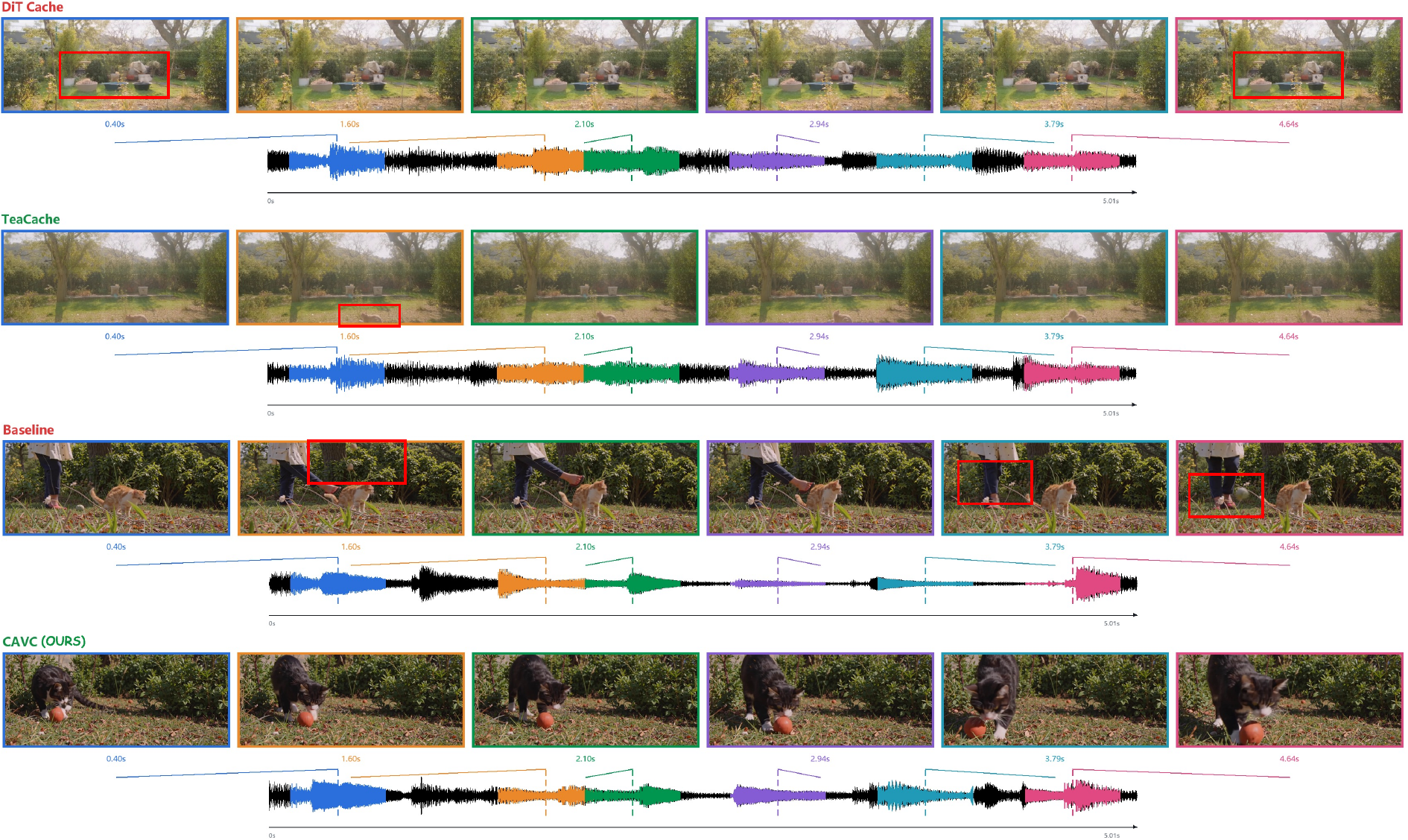}
    
    \caption{Qualitative comparison of audio-visual joint generation. Given the text prompt ``a cat playing with a ball in the yard'', the generic Cache-DiT and TeaCache baselines exhibit visible artifacts and weaker prompt following. For this 121-frame 1K example, their cache-state estimation and update operations also introduce net overhead (see Table~\ref{tab:visual_pixel_benchmarks}). In contrast, CAVC achieves a 1.15$\times$ inference speedup while preserving stronger visual quality, spatio-temporal consistency, and dynamics.}
    \label{fig:qualitative_cat}
\end{figure*}

\noindent\textbf{Stage 1 (Phase B): Video-Only Continuation with Cached Audio.}
Let $t_{\mathrm{ref}}$ denote the first timestep of terminal joint refinement, with $t_{\mathrm{ref}}=3$ in our default setting. During the cached interval $\mathcal{I}=\{\tau-1,\tau-2,\ldots,t_{\mathrm{ref}}+1\}$, CAVC simultaneously bypasses the audio denoiser and AV cross-modal attention. The audio latent continues its trajectory using the cached audio denoiser output:
\begin{equation}
    \hat{y}_a^t := \hat{y}_a^*, \qquad
    \tilde{a}_{t-1} = S_t^a\!\left(\tilde{a}_t,\, \hat{y}_a^*\right).
\end{equation}
Critically, CAVC caches the denoiser output rather than freezing the latent state directly, enabling the scheduler to continue smooth variance integration along the original noise schedule $\sigma_t$. The validity of this cross-noise-level output reuse is discussed in Supplement~\ref{app:reuse}. Simultaneously, the video branch proceeds independently:

\begin{equation}
    \tilde{y}_v^t = g_v\!\left(\tilde{v}_t, c, \sigma_t\right), \qquad
    \tilde{v}_{t-1} = S_t^v\!\left(\tilde{v}_t,\, \tilde{y}_v^t\right),
\end{equation}
where $g_v$ denotes the video denoiser with AV cross-attention modules bypassed via residual skip connections while fully preserving the video self-/temporal attention and text conditioning. This eliminates the computationally expensive $O(N_vN_a)$ cross-attention and prevents cached audio features from over-constraining video dynamics.

\noindent\textbf{Stage 2 (Phase C): Final Joint Refinement.}
CAVC restores full joint denoising for $t=t_{\mathrm{ref}},\ldots,1$, producing the final states at $t=0$:
\begin{align}
    \tilde{y}_v^t &= f_v(\tilde{v}_t, \tilde{a}_t, c, \sigma_t), &
    \tilde{y}_a^t &= f_a(\tilde{v}_t, \tilde{a}_t, c, \sigma_t), \\
    \tilde{v}_{t-1} &= S_t^v(\tilde{v}_t,\tilde{y}_v^t), &
    \tilde{a}_{t-1} &= S_t^a(\tilde{a}_t,\tilde{y}_a^t).
\end{align}
This lightweight recoupling empirically reduces cross-modal discrepancies from Phase B and improves audio-visual synchronization before decoding. Table~\ref{tab:recoverability} evaluates this asymmetric recoverability.

\newcommand{\gain}[1]{\textcolor{green!50!black}{\scriptsize{~(+#1)}}}
\newcommand{\drop}[1]{\textcolor{gray}{\scriptsize{~(#1)}}}
\newcommand{\impdown}[1]{\textcolor{green!50!black}{\scriptsize{~(#1)}}} 
\newcommand{\fail}[1]{\textcolor{gray}{\scriptsize{~(#1)}}}
\newcommand{\recovery}[1]{\textcolor{green!50!black}{\scriptsize{~(#1\% Rec.)}}}
\newcommand{\norecovery}[1]{\textcolor{gray}{\scriptsize{~(#1\% Rec.)}}}

\begin{table*}[t]
\centering
\caption{Evaluation of \textbf{Visual Quality} and \textbf{Low-Level Statistics} across three benchmarks. Low-level statistics are no-reference diagnostic metrics; definitions are provided in the supplement. $\uparrow$ indicates higher is better, and $\downarrow$ indicates lower is better. \textbf{Bold} indicates the best performance. CAVC provides a favorable quality-efficiency trade-off while achieving inference speedups.}

\label{tab:visual_pixel_benchmarks}
\resizebox{0.96\linewidth}{!}{
\begin{tabular}{@{}llc | ccccc | ccc@{}}
\toprule
\multirow{2}{*}{\textbf{Benchmark}} & \multirow{2}{*}{\textbf{Method}} & \multirow{2}{*}{\textbf{Speedup $\uparrow$}} & \multicolumn{5}{c|}{\textbf{Visual Quality} \cite{huang2025vbench++}} & \multicolumn{3}{c}{\textbf{Low-Level Statistics}} \\
\cmidrule(lr){4-8} \cmidrule(l){9-11}
& & & \textbf{Subj.Con$\uparrow$} & \textbf{Bg.Con$\uparrow$} & \textbf{Mot.Smth$\uparrow$} & \textbf{Dyn.Deg$\uparrow$} & \textbf{Aes.Qual$\uparrow$} & \textbf{Natural$\uparrow$} & \textbf{Temporal$\uparrow$} & \textbf{Flicker$\downarrow$} \\
\midrule
\multirow{4}{*}{AVGen-Bench} 
& TeaCache \cite{teacache}  & 0.45$\times$ & 0.864 & 0.920 & 0.989 & 0.272 & 0.452 & 0.473 & 0.619 & 0.522 \\
& Cache-DiT \cite{cache-dit} & 0.24$\times$ & 0.869 & 0.920 & 0.989 & 0.298 & 0.462 & 0.463 & 0.612 & 0.517 \\
\cmidrule{2-11}
& Full AV \cite{hacohen2026ltx2efficientjointaudiovisual}   & 1.00$\times$ & 0.7895 & 0.8622 & 0.9200 & 0.655 & \textbf{0.524} & 0.5272 & 0.6534 & 0.3346 \\
\rowcolor{gray!10} & CAVC (Ours)      & \textbf{1.10$\times$}\gain{10.0\%} & \textbf{0.874}\gain{10.7\%} & \textbf{0.925}\gain{7.3\%} & \textbf{0.991}\gain{7.7\%} & \textbf{0.6701}\gain{2.3\%} & 0.522 & \textbf{0.5540}\gain{5.1\%} & \textbf{0.6600} & \textbf{0.2800}\impdown{-16.3\%} \\
\midrule
\multirow{4}{*}{VABench T2AV} 
& TeaCache \cite{teacache}  & 0.64$\times$ & 0.936 & 0.956 & 0.993 & 0.099 & 0.482 & 0.409 & 0.668 & 0.180 \\
& Cache-DiT \cite{cache-dit} & 0.35$\times$ & \textbf{0.942} & \textbf{0.959} & 0.994 & 0.094 & 0.500 & 0.402 & 0.655 & 0.170 \\
\cmidrule{2-11}
& Full AV \cite{hacohen2026ltx2efficientjointaudiovisual}   & 1.00$\times$ & 0.9280 & 0.9530 & 0.994 & 0.4320 & 0.5540 & 0.4193 & 0.6621 & \textbf{0.1608} \\
\rowcolor{gray!10} & CAVC (Ours)      & \textbf{1.35$\times$}\gain{34.7\%} & 0.932 & 0.954 & \textbf{0.994} & \textbf{0.4470}\gain{3.5\%} & \textbf{0.559} & \textbf{0.509}\gain{21.4\%} & \textbf{0.718}\gain{8.4\%} & 0.176 \\
\midrule
\multirow{4}{*}{General-80} 
& TeaCache \cite{teacache}  & 0.39$\times$ & 0.932 & 0.957 & 0.990 & 0.275 & 0.519 & 0.477 & 0.667 & 0.467 \\
& Cache-DiT \cite{cache-dit} & 0.20$\times$ & 0.938 & 0.961 & 0.991 & 0.237 & 0.544 & 0.520 & 0.649 & 0.493 \\
\cmidrule{2-11}
& Full AV \cite{hacohen2026ltx2efficientjointaudiovisual}   & 1.00$\times$ & 0.9460 & 0.9540 & 0.9920 & 0.3800 & 0.5560 & \textbf{0.6349} & 0.7136 & 0.3346 \\
\rowcolor{gray!10} & CAVC (Ours)      & \textbf{1.20$\times$}\gain{20.0\%} & \textbf{0.949} & \textbf{0.961} & \textbf{0.993} & \textbf{0.512}\gain{34.7\%} & \textbf{0.561} & 0.490 & \textbf{0.718} & \textbf{0.263}\impdown{-21.4\%} \\
\bottomrule
\end{tabular}
}

\end{table*}

\begin{table*}[t]
\centering
\caption{Evaluation of \textbf{Automated Audio \& MLLM-Based Quality} and \textbf{Audio-Visual Alignment \& Physics} across three benchmarks. AV-F1 is an onset-based diagnostic metric. $\uparrow$ indicates higher is better; \textbf{bold} indicates best performance.}

\label{tab:audio_physics_benchmarks}
\resizebox{0.96\linewidth}{!}{
\begin{tabular}{@{}llc | ccccc | ccc@{}}
\toprule
\multirow{2}{*}{\textbf{Benchmark}} & \multirow{2}{*}{\textbf{Method}} & \multirow{2}{*}{\textbf{Speedup $\uparrow$}} & \multicolumn{5}{c|}{\textbf{Automated Audio \& MLLM Quality}} & \multicolumn{3}{c}{\textbf{Audio-Visual \& Physics}} \\
\cmidrule(lr){4-8} \cmidrule(l){9-11}
& & & \textbf{DNSMOS$\uparrow$} & \textbf{CLAP$\uparrow$} & \textbf{Aud.Real$\uparrow$} & \textbf{Expr.$\uparrow$} & \textbf{Vis.Real$\uparrow$} & \textbf{AV-F1$\uparrow$} & \textbf{AV Sync$\uparrow$} & \textbf{Lo-Phy$\uparrow$} \\
\midrule
\multirow{4}{*}{AVGen-Bench} 
& TeaCache \cite{teacache}  & 0.45$\times$ & 1.607 & 0.366 & 4.055 & 4.264 & 4.609 & 0.601 & 0.429 & \textbf{3.936} \\
& Cache-DiT \cite{cache-dit} & 0.24$\times$ & 1.676 & 0.418 & 4.068 & \textbf{4.319} & 4.566 & \textbf{0.614} & 0.249 & 3.911 \\
\cmidrule{2-11}
& Full AV \cite{hacohen2026ltx2efficientjointaudiovisual}   & 1.00$\times$ & \textbf{1.706} & 0.435 & 3.925 & 4.070 & 4.475 & 0.5381 & \textbf{0.561} & 3.847 \\
\rowcolor{gray!10} & CAVC (Ours)      & \textbf{1.10$\times$}\gain{10.0\%} & 1.688\drop{-1.1\%} & \textbf{0.438} & \textbf{4.077}\gain{3.9\%} & 4.294\gain{5.5\%} & \textbf{4.723}\gain{5.5\%} & 0.5650\gain{5.0\%} & 0.525\drop{-6.4\%} & 3.872 \\
\midrule
\multirow{4}{*}{VABench T2AV} 
& TeaCache \cite{teacache}  & 0.64$\times$ & 1.324 & 0.294 & 3.912 & \textbf{4.392} & 4.468 & \textbf{0.659} & 0.549 & \textbf{4.503} \\
& Cache-DiT \cite{cache-dit} & 0.35$\times$ & 1.399 & \textbf{0.366} & 3.918 & 4.386 & \textbf{4.497} & 0.652 & 0.470 & 4.455 \\
\cmidrule{2-11}
& Full AV \cite{hacohen2026ltx2efficientjointaudiovisual}   & 1.00$\times$ & 1.383 & 0.359 & 3.678 & 4.072 & 4.309 & 0.5150 & \textbf{0.608} & 4.255 \\
\rowcolor{gray!10} & CAVC (Ours)      & \textbf{1.35$\times$}\gain{34.7\%} & \textbf{1.498}\gain{8.3\%} & 0.359 & \textbf{3.965}\gain{7.8\%} & 4.333\gain{6.4\%} & 4.491\gain{4.2\%} & 0.552\gain{7.2\%} & 0.594\drop{-2.3\%} & 4.231 \\
\midrule
\multirow{4}{*}{General-80} 
& TeaCache \cite{teacache}  & 0.39$\times$ & 1.197 & 0.313 & 4.188 & \textbf{4.575} & \textbf{4.812} & \textbf{0.688} & 0.563 & \textbf{4.350} \\
& Cache-DiT \cite{cache-dit} & 0.20$\times$ & 1.270 & \textbf{0.379} & \textbf{4.225} & 4.562 & \textbf{4.812} & 0.682 & 0.493 & 4.263 \\
\cmidrule{2-11}
& Full AV \cite{hacohen2026ltx2efficientjointaudiovisual}   & 1.00$\times$ & \textbf{1.306} & 0.284 & 4.000 & 4.160 & 4.620 & 0.5381 & 0.496 & 4.180 \\
\rowcolor{gray!10} & CAVC (Ours)      & \textbf{1.20$\times$}\gain{20.0\%} & 1.283\drop{-1.8\%} & 0.324\gain{14.1\%} & 4.125\gain{3.1\%} & 4.450\gain{7.0\%} & 4.650 & 0.644\gain{19.7\%} & \textbf{0.563}\gain{13.5\%} & 4.300\gain{2.9\%} \\
\bottomrule
\end{tabular}
}
\end{table*}

\subsection{Theoretical Analysis}
\label{sec:theory}

We formalize the deviation between CAVC and the full sampler, characterizing how the approximation errors introduced during Phase B accumulate under standard smoothness assumptions. This analysis should be understood as a perturbation characterization rather than a claim of perfect recovery.

\noindent\textbf{Error Sources.}
Post-triggering, CAVC introduces two approximation errors, empirically quantified in Figure~\ref{fig:convergence_errors}. The audio cache error $\epsilon_a$ captures the discrepancy of freezing the audio prediction:
\begin{equation}
    \left\|f_a(v_t, a_t, c, \sigma_t) - \hat{y}_a^*\right\| \leq \epsilon_a,
\end{equation}
and the \emph{video-only residual} $\epsilon_v$ captures the impact of discarding AV cross-attention:
\begin{equation}
    \left\|f_v(v_t, a_t, c, \sigma_t) - g_v(v_t, c, \sigma_t)\right\| \leq \epsilon_v.
\end{equation}
The adaptive threshold is an empirical proxy for entering a cache-tolerant audio regime, while Phase A warm-up reduces the dominant cross-modal dependency before AV attention is bypassed.

\noindent\textbf{Latent Deviation Bound.}
Assuming the scheduler satisfies local Lipschitz conditions with constants $\alpha_t$ (w.r.t.\ the latent) and $\beta_t$ (w.r.t.\ the prediction), the cumulative latent deviations $e_a^t = \|a_t - \tilde{a}_t\|$ and $e_v^t = \|v_t - \tilde{v}_t\|$ within the cached interval $\mathcal{I} = \{\tau-1,\tau-2,\ldots,t_{\mathrm{ref}}+1\}$ satisfy:
\begin{align}
    e_a^{t-1} &\leq \alpha_t\, e_a^t + \beta_t\, \epsilon_a, \\
    e_v^{t-1} &\leq (\alpha_t + \beta_t L_g)\, e_v^t + \beta_t\, \epsilon_v,
\end{align}
where $L_g$ is the local Lipschitz constant of $g_v$. Because Phase A includes the full update at trigger step $\tau$, the two trajectories agree at the start of Phase B, i.e., $e_a^{\tau-1}=e_v^{\tau-1}=0$. Unrolling over $\mathcal{I}$ yields:
\begin{align}
    e_a^{t_{\mathrm{ref}}} &\leq \sum_{t=t_{\mathrm{ref}}+1}^{\tau-1}
    \left(\prod_{j=t_{\mathrm{ref}}+1}^{t-1} \alpha_j\right) \beta_t\, \epsilon_a, \\
    e_v^{t_{\mathrm{ref}}} &\leq \sum_{t=t_{\mathrm{ref}}+1}^{\tau-1}
    \left(\prod_{j=t_{\mathrm{ref}}+1}^{t-1}(\alpha_j + \beta_j L_g)\right) \beta_t\, \epsilon_v.
\end{align}
These bounds characterize how terminal deviation depends on $\epsilon_a$, $\epsilon_v$, and caching duration $|\mathcal{I}|$ under the stated smoothness assumptions.

\noindent\textbf{Output Quality Bound.}
Let $E_t=e_a^t+e_v^t$. Assume that each terminal joint update in Phase C is locally Lipschitz in this product-state norm with constant $\kappa_t$, and define $K_C=\prod_{t\in\mathcal{J}}\kappa_t$ for $\mathcal{J}=\{t_{\mathrm{ref}},t_{\mathrm{ref}}-1,\ldots,1\}$. Let decoders $D_v$, $D_a$, and a smooth scalar output functional $Q$ be locally Lipschitz with constants $L_D^v$, $L_D^a$, and $L_Q$. The final output discrepancy satisfies:
\begin{equation}
\begin{aligned}
    \left|Q(y)-Q(\tilde{y})\right|
    &\leq L_Q\!\left(L_D^v e_v^0+L_D^a e_a^0\right) \\
    &\leq L_Q\max\!\left\{L_D^v,L_D^a\right\}
    K_C\!\left(e_v^{t_{\mathrm{ref}}}+e_a^{t_{\mathrm{ref}}}\right).
\end{aligned}
\end{equation}
This conditional bound characterizes error propagation through Phase C; it does not require the terminal updates to be contractive. Their empirical recovery behavior is evaluated in Table~\ref{tab:recoverability}. The complete derivation, including nested state bounding and multi-step accumulation, is detailed in Supplement~\ref{sec:appendix_derivation}.

\subsection{Complexity Analysis}
\label{sec:complexity}

Let $N_1$ and $N_2$ denote the numbers of Stage-1 steps and terminal Phase-C steps, respectively, with $N_2=3$ in the native LTX-2.3 pipeline. Let $k_\tau=T-\tau$ be the zero-based trigger index and $n_\tau=k_\tau+1$ the number of full joint Stage-1 evaluations executed through the trigger step. Thus, unlike the diffusion timestep $\tau$, $n_\tau$ is a step count. Let $C_v^{(s)}$, $C_a^{(s)}$, and $C_{av}^{(s)}$ denote the per-step costs of the video branch, audio branch, and cross-attention in stage $s\in\{1,2\}$; $C_d$ is the per-step detection cost, and $C_{\mathrm{fix}}$ collects shared stage-transition, decoding, and I/O costs. The full sampler costs:
\begin{equation}
\begin{aligned}
    \mathcal{C}_{\mathrm{full}}
    ={}&N_1\!\left(C_v^{(1)}+C_a^{(1)}+C_{av}^{(1)}\right) \\
    &+N_2\!\left(C_v^{(2)}+C_a^{(2)}+C_{av}^{(2)}\right)
    +C_{\mathrm{fix}}.
\end{aligned}
\end{equation}
During Stage 1, CAVC executes the audio branch and cross-attention only before caching, while leaving the native terminal Phase C unchanged:
\begin{equation}
\begin{aligned}
    \mathcal{C}_{\mathrm{CAVC}}
    ={}&N_1 C_v^{(1)}
    +n_\tau\!\left(C_a^{(1)}+C_{av}^{(1)}\right)
    +n_\tau C_d \\
    &+N_2\!\left(C_v^{(2)}+C_a^{(2)}+C_{av}^{(2)}\right)
    +C_{\mathrm{fix}}.
\end{aligned}
\end{equation}
Since $C_d\ll C_a^{(1)}+C_{av}^{(1)}$, the saving is $\Delta\mathcal{C}=(N_1-n_\tau)(C_a^{(1)}+C_{av}^{(1)})-n_\tau C_d$, yielding speedup:
\begin{equation}
    r=\frac{
    \begin{aligned}
        &N_1(C_v^{(1)}+C_a^{(1)}+C_{av}^{(1)}) \\
        &\quad+N_2(C_v^{(2)}+C_a^{(2)}+C_{av}^{(2)})+C_{\mathrm{fix}}
    \end{aligned}
    }{
    \begin{aligned}
        &N_1C_v^{(1)}+n_\tau(C_a^{(1)}+C_{av}^{(1)})+n_\tau C_d \\
        &\quad+N_2(C_v^{(2)}+C_a^{(2)}+C_{av}^{(2)})+C_{\mathrm{fix}}
    \end{aligned}
    }.
\end{equation}
The shared Phase-C and fixed costs remain in both the numerator and denominator, explaining why the measured wall-clock gain is smaller than the Stage-1 branch-level saving. CAVC empirically achieves 1.10--1.35$\times$ wall-clock acceleration while leaving terminal refinement unchanged.

\section{Experiments}

\subsection{Experimental Setup}

\textbf{Benchmarks.} We evaluate on AVGen-Bench~\cite{avgenbench}, VABench T2AV~\cite{hua2025vabench}, and General-80. Together, they span general scenes and audio-visual synchronization challenges.

\noindent\textbf{Implementation Details \& Baselines.} CAVC is implemented on LTX-2.3-22B-dev \cite{hacohen2026ltx2efficientjointaudiovisual} with the official distilled LoRA. We compare against Full AV, Cache-DiT, and TeaCache under matched generation specifications on one NVIDIA H20. Complete settings and evaluation protocols are provided in the supplement.

\begin{table*}[t]
\centering
\caption{Comprehensive ablation studies. (a) Branch caching/freezing strategy. (b) Alternative caching strategies. (c) Phase processing components and terminal refinement. (d) Hyperparameter analysis on threshold ($\theta$) and window size ($W$). The configuration $\theta=0.30, W=3$ is selected as the default setting, balancing acceleration and recoverability. Audio@ denotes the average forward execution index $k_\tau$ at which audio caching is triggered, not the decreasing diffusion timestep $\tau$. $\uparrow$ indicates higher is better, and $\downarrow$ indicates lower is better.}
\label{tab:ablation_master}

\scalebox{0.92}{
\begin{minipage}{\textwidth} 

\begin{minipage}[t]{0.49\linewidth}

    \begin{subtable}{\linewidth}
    \centering
    \caption{Branch Freezing Strategy}
    \label{tab:ablation_branch}
    \renewcommand{\arraystretch}{1.35}
    \resizebox{\linewidth}{!}{
    \begin{tabular}{@{}l | c | cc | cc@{}}
    \toprule
    \multirow{2}{*}{\textbf{Method (Freeze)}} & \multirow{2}{*}{\textbf{Speed$\uparrow$}} & \multicolumn{2}{c|}{\textbf{Video Quality}} & \multicolumn{2}{c}{\textbf{Audio \& Align}} \\
    \cmidrule(lr){3-4} \cmidrule(l){5-6}
    & & \textbf{Bg.$\uparrow$} & \textbf{Dyn.$\uparrow$} & \textbf{AV-F1$\uparrow$} & \textbf{CLAP$\uparrow$} \\
    \midrule
    Video Only        & 2.84$\times$ & 0.950 & 0.460 & 0.593 & \textbf{0.362} \\
    Audio \& Video    & \textbf{3.01$\times$} & 0.949 & 0.440 & \textbf{0.640} & 0.316 \\
    Audio Only (Ours) & 1.19$\times$ & \textbf{0.960}\gain{1.2\%} & \textbf{0.700}\gain{59.1\%} & 0.634 & 0.312 \\
    \bottomrule
    \end{tabular}
    }
    \end{subtable}

    \vspace{1em} 
    
    \vspace*{10pt} 
    
    \begin{subtable}{\linewidth}
    \centering
    \caption{Alternative Caching Strategies}
    \label{tab:ablation_cache}
    \renewcommand{\arraystretch}{1.35}
    \resizebox{\linewidth}{!}{
    \begin{tabular}{@{}l | c | ccc | cc@{}}
    \toprule
    \multirow{2}{*}{\textbf{Method}} & \multirow{2}{*}{\textbf{Speed$\uparrow$}} & \multicolumn{3}{c|}{\textbf{Video Quality}} & \multicolumn{2}{c}{\textbf{Audio \& Align}} \\
    \cmidrule(lr){3-5} \cmidrule(l){6-7}
    & & \textbf{Bg.$\uparrow$} & \textbf{Dyn.$\uparrow$} & \textbf{Aes.$\uparrow$} & \textbf{AV-F1$\uparrow$} & \textbf{CLAP$\uparrow$} \\
    \midrule
    Baseline             & 1.00$\times$ & 0.958 & 0.440 & 0.562 & 0.617 & 0.287 \\
    Cache Only           & 1.00$\times$ & 0.960 & 0.440 & 0.566 & 0.641 & 0.291 \\
    Fixed 20\%           & 1.09$\times$ & 0.960 & 0.500 & 0.563 & 0.643 & \textbf{0.304} \\
    Fixed 80\%           & 1.01$\times$ & 0.959 & 0.460 & 0.560 & 0.618 & 0.290 \\
    Dynamic (Ours)       & \textbf{1.19$\times$}\gain{19\%} & \textbf{0.963} & \textbf{0.560}\gain{27\%} & \textbf{0.572}\gain{1.8\%} & \textbf{0.674}\gain{9.2\%} & 0.301 \\
    \bottomrule
    \end{tabular}
    }
    \end{subtable}

\end{minipage}
\hfill
\begin{minipage}[t]{0.46\linewidth}

    \begin{subtable}{\linewidth}
    \centering
    \caption{Stage Processing Ablation}
    \label{tab:ablation_stage}
    \renewcommand{\arraystretch}{1.1}
    \resizebox{0.85\linewidth}{!}{
    \begin{tabular}{@{}l | ccc | c@{}}
    \toprule
    \multirow{2}{*}{\textbf{Method}} & \multicolumn{3}{c|}{\textbf{Video Quality (VBench)}} & \textbf{Align} \\
    \cmidrule(lr){2-4} \cmidrule(l){5-5}
    & \textbf{Subj.$\uparrow$} & \textbf{Smooth$\uparrow$} & \textbf{Dyn.$\uparrow$} & \textbf{Mot.$\uparrow$} \\
    \midrule
    Stage 1 (Base)     & 0.936 & 0.989 & 0.540 & 1.140 \\
    Stage 1 (Ours)     & 0.938 & 0.989 & 0.580\gain{7.4\%} & 1.199 \\
    Full Stages (Ours) & \textbf{0.953}\gain{1.8\%} & \textbf{0.993} & \textbf{0.620}\gain{14.8\%} & \textbf{1.283}\gain{12.5\%} \\
    \bottomrule
    \end{tabular}
    }
    \end{subtable}

    \vspace{1.3em} 

    \begin{subtable}{\linewidth}
    \centering
    \caption{Hyperparameter Analysis ($\theta$ \& $W$). The default setting $\theta=0.30, W=3$ provides a balanced trade-off between trigger timing and inference acceleration.}
    \label{tab:ablation_hyperparam}
    \renewcommand{\arraystretch}{0.9}
    \resizebox{0.84\linewidth}{!}{
    \begin{tabular}{@{}cc | ccc@{}}
    \toprule
    \textbf{Threshold ($\theta$)} & \textbf{Window ($W$)} & \textbf{Time(s)$\downarrow$} & \textbf{Speedup$\uparrow$} & \textbf{Audio@} \\
    \midrule
    \multicolumn{2}{@{}c|}{\textit{Baseline}} & $217.2 \pm 0.7$ & 1.00$\times$ & - \\
    \midrule
    0.15 & 3 & $192.8 \pm 4.5$ & 1.13$\times$ & 14.9 \\
    0.30 & 1 & $181.4 \pm 1.0$ & 1.20$\times$ & 1.5 \\
    0.30 & 2 & $182.4 \pm 1.2$ & 1.19$\times$ & 2.9 \\
    \rowcolor{gray!15}
    \textbf{0.30} & \textbf{3} & $\mathbf{184.8 \pm 2.9}$ & \textbf{1.18$\times$} & \textbf{5.5} \\
    0.50 & 1 & $180.6 \pm 0.6$ & 1.20$\times$ & 1.1 \\
    0.50 & 2 & $181.4 \pm 1.2$ & 1.20$\times$ & 2.3 \\
    0.50 & 3 & $182.9 \pm 0.9$ & 1.19$\times$ & 3.3 \\
    0.80 & 1 & $180.4 \pm 0.7$ & 1.20$\times$ & 1.0 \\
    0.80 & 2 & $181.2 \pm 0.6$ & 1.20$\times$ & 2.0 \\
    0.80 & 3 & $182.5 \pm 0.9$ & 1.19$\times$ & 3.0 \\
    \bottomrule
    \end{tabular}
    }
    \end{subtable}

\end{minipage}

\end{minipage}
} 
\end{table*}

\subsection{Main Results: Quality-Efficiency Trade-off}

Tables~\ref{tab:visual_pixel_benchmarks} and~\ref{tab:audio_physics_benchmarks} detail benchmark comparisons. Metric formulations and evaluator configurations are provided in Supplement~\ref{sec:appendix_metrics} to ensure full reproducibility.

\noindent\textbf{Inference Acceleration.}
By bypassing mid-trajectory audio computation and AV attention, CAVC achieves \textbf{1.10--1.35$\times$} wall-clock speedups, completely training-free. Speedups vary by prompt: stable audio patterns trigger caching early, while complex multi-event audio delays it, allowing CAVC to gracefully approach the full sampler.

\noindent\textbf{Unlocking Visual Dynamics.}
Removing mid-stage AV over-constraints allows the video branch to evolve without being anchored by static audio features, improving several motion and temporal statistics. \textit{Dynamic Degree} increases by \textbf{+34.7\%} (General-80), and \textit{Flicker} drops by \textbf{-16.3\%} (AVGen-Bench). Stable or improved onset-based AV-F1 suggests that Phase C re-couples the modalities after caching.

\noindent\textbf{Preserving Audio Fidelity and Alignment.}
Audio caching incurs limited acoustic deviations (DNSMOS drops $-1.1\%$ on AVGen-Bench), while onset-based AV-F1 improves (\textbf{+5.0\%--+19.7\%}) across datasets. Fine-grained AV Sync shows minor residual drops on some benchmarks, indicating that terminal refinement substantially reduces, but does not always eliminate, the accumulated cross-modal drift.

\subsection{Ablation Studies}

Table~\ref{tab:ablation_master} isolates design components; Table~\ref{tab:recoverability} validates our modality-asymmetric recoverability premise.

\noindent\textbf{Validation of Asymmetric Recoverability (Table~\ref{tab:recoverability}).}
Terminal recoupling directly restores the audio semantic capability lost during cached continuation: across the evaluated refinements, 81.5--91.3\% of the CLAP gap to Full AV is closed. In contrast, Dynamics recovery after video-side caching ranges from $-16.7\%$ to 0.0\%, showing that terminal recoupling does not repair the lost motion capability. This branch-dependent recovery behavior supports allocating the approximation to the audio side while strictly preserving video refinement.

\begin{table}[htbp]
\centering
\caption{\textbf{Empirical validation of modality-asymmetric capability recovery.} Percentages denote gap closure from the corresponding 0-step cached condition toward Full AV, measured by CLAP for audio capability and Dynamics for video capability. Terminal recoupling repairs most audio capability lost to caching, whereas video dynamics remain unrecovered.}
\label{tab:recoverability}

\resizebox{0.96\linewidth}{!}{
\begin{tabular}{@{}llcccc@{}}
\toprule
\textbf{Cached Branch} & \textbf{Phase C} & \textbf{Dyn.$\uparrow$} & \textbf{Aes.Qual$\uparrow$} & \textbf{AV Sync$\uparrow$} & \textbf{CLAP$\uparrow$} \\
\midrule
None (Full AV) & 3 steps & 0.600 & 0.602 & 0.511 & 0.271 \\
\midrule
Audio (CAVC)   & 0 steps & 0.560 & 0.563 & 0.495 & -0.005\norecovery{0.0} \\
Audio (CAVC)   & 1 step  & 0.640 & 0.603 & 0.679 & 0.220\recovery{81.5} \\
Audio (CAVC)   & 3 steps & 0.620 & 0.596 & 0.502 & 0.235\recovery{87.0} \\
Audio (CAVC)   & 5 steps & 0.620 & 0.596 & 0.566 & 0.247\recovery{91.3} \\
\midrule
Video          & 0 steps & 0.360\norecovery{0.0} & 0.286 & 0.290 & 0.300 \\
Video          & 1 step  & 0.320\norecovery{-16.7} & 0.588 & 0.590 & 0.237 \\
Video          & 3 steps & 0.360\norecovery{0.0} & 0.575 & 0.456 & 0.289 \\
Video          & 5 steps & 0.360\norecovery{0.0} & 0.570 & 0.464 & 0.294 \\
\bottomrule
\end{tabular}
}
\end{table}

\noindent\textbf{Why Asymmetric Caching? (Table~\ref{tab:ablation_master}a).}
Freezing the video branch or both branches yields ${\sim}3\times$ speedups but severely degrades video dynamics (Dyn.\ drops to 0.460/0.440), as interrupting continuous spatio-temporal self-attention disrupts complex motion synthesis. Audio-only caching preserves this video refinement, driving dynamics to 0.700 (+59.1\%). This result supports allocating the approximation to the audio side while preserving video computation.

\noindent\textbf{Effect of Phase C Refinement (Table~\ref{tab:ablation_master}c).}
While Phase B independently improves video dynamics (Dyn.\ 0.540$\to$0.580), the full sandwich structure further improves motion alignment (1.199$\to$1.283, +6.7\%) after terminal joint denoising. This supports Phase C as an effective refinement mechanism for reducing the drift introduced during Phase B.

\noindent\textbf{Dynamic vs.\ Fixed Triggering.}
As shown in Table~\ref{tab:ablation_master}b, fixed schedules fail to balance quality and efficiency: \textit{Fixed 80\%} yields negligible speedup ($1.01\times$), while \textit{Fixed 20\%} decouples before foundational structures form. In contrast, our adaptive trigger provides a stronger operating point for both speed ($1.19\times$) and onset-based AV-F1 (0.674).

\noindent\textbf{Hyperparameter Robustness (Table~\ref{tab:ablation_master}d).}
The threshold $\theta$ and window size $W$ determine the audio-cache trigger. A stricter threshold ($\theta=0.15$) delays caching (Audio@ $k_\tau=14.9$) and limits speedup to $1.13\times$, whereas looser thresholds trigger earlier with only marginal additional acceleration. We choose $\theta=0.30, W=3$ as the default, which activates caching at a moderate forward execution index (Audio@ $k_\tau=5.5$) and achieves a stable $1.18\times$ speedup.

\section{Conclusion}

We present Convergence-Adaptive Audio-Visual Cache (CAVC), a training-free inference acceleration framework for asymmetric dual-stream audio-visual diffusion transformers. CAVC is grounded in modality-asymmetric recoverability: audio-side drift introduced by mid-stage caching can be substantially reduced by a few terminal joint denoising steps, whereas video-side degradation from interrupted refinement is difficult to recover. Guided by this insight, CAVC preserves the full video branch, caches recoverable audio-side computation, bypasses AV attention in the middle stage, and restores brief joint denoising at the end. Experiments show that CAVC achieves 1.10--1.35$\times$ wall-clock acceleration, improves several visual dynamics and temporal statistics, and maintains competitive audio fidelity and AV alignment.

\noindent\textbf{Limitations.} By preserving the video backbone to ensure visual fidelity, CAVC's maximum speedup is inherently bounded by video-side computation. For continuously shifting acoustic events, the adaptive trigger delays caching, causing efficiency to approach the baseline sampler. Minor AV synchronization offsets in rapid multi-event scenes can often be reduced by increasing Phase C steps at marginal cost.

\bibliographystyle{aaai2027}
\bibliography{aaai2026}

\clearpage
\appendix
\section{Detailed Experimental Protocol}
\label{app:protocol}

\subsection{Generation Configuration}
Table~\ref{tab:app_generation_config} records the generation settings used for the reported experiments. Each sample is generated independently on one NVIDIA H20 GPU with 96\,GB memory. CPU offload is enabled for all four methods in the reported runs.

\begin{table}[htbp]
\centering
\caption{Generation configuration shared by Full AV, CAVC, Cache-DiT, and TeaCache.}
\label{tab:app_generation_config}
\small
\resizebox{\linewidth}{!}{
\begin{tabular}{@{}ll@{}}
\toprule
\textbf{Item} & \textbf{Setting} \\
\midrule
Base model & LTX-2.3-22B-dev \\
Distilled LoRA & \texttt{ltx-2.3-22b-distilled-lora-384-1.1}, strength 0.8 \\
Spatial upscaler & \texttt{ltx-2.3-spatial-upscaler-x2-1.1} \\
Denoising steps / generation seed & 30 / 100 \\
AVGen-Bench and VABench & $1280\!\times\!704$, 241 frames, 24 FPS \\
General-80 & $1536\!\times\!1024$, 121 frames, 24 FPS \\
Video CFG / Audio CFG & 3.0 / 7.0 \\
Video STG / Audio STG & 1.0 / 1.0 \\
Video/Audio modality scale & 3.0 / 3.0 \\
Video/Audio guidance rescale & 0.7 / 0.7 \\
STG block & 28 \\
Numerical precision & BF16 \\
Hardware & One NVIDIA H20 (96\,GB) per generated sample \\
Batch size / generation concurrency & 1 / 1 \\
CPU offload & Enabled for all reported methods \\
\bottomrule
\end{tabular}
}
\end{table}

\subsection{Acceleration-Method Configuration}
The behavior-affecting cache parameters are listed in Table~\ref{tab:app_cache_config}. Cache-DiT uses its official LTX2 implementation. TeaCache uses a public implementation adapted to the LTX-2.3 dual-stream pipeline. The exact frozen repository snapshots and the TeaCache adaptation patch are recorded in the manifest distributed with the anonymous code artifact; no floating repository branch is used for the reported runs.

\begin{table}[htbp]
\centering
\caption{Method-specific cache configuration. A value of $-1$ means that no explicit maximum is imposed.}
\label{tab:app_cache_config}
\small
\resizebox{\linewidth}{!}{
\begin{tabular}{@{}lll@{}}
\toprule
\textbf{Method} & \textbf{Parameter} & \textbf{Value} \\
\midrule
CAVC & convergence threshold $\theta$ / window $W$ & 0.30 / 3 \\
\midrule
Cache-DiT & warm-up steps & 8 \\
Cache-DiT & front/back compute blocks & 8 / 0 \\
Cache-DiT & residual-difference threshold & 0.08 \\
Cache-DiT & max cached / continuous cached steps & $-1$ / $-1$ \\
Cache-DiT & TaylorSeer / transformer compilation & disabled / disabled \\
\midrule
TeaCache & \texttt{rel\_l1\_thresh} & 0.20 \\
TeaCache & active interval & 0.15--1.00 of the denoising trajectory \\
TeaCache & polynomial coefficients & [21.4700694, $-12.8016453$, 2.31279151, 0.792487521, 0.00969274326] \\
\bottomrule
\end{tabular}
}
\end{table}

For TeaCache, the relative first-block change is
\begin{equation}
 d_k=\frac{\operatorname{mean}(|x^{(k)}-x^{(k-1)}|)}
 {\max(\operatorname{mean}(|x^{(k-1)}|),10^{-8})}.
\end{equation}
The polynomial mapping of $d_t$ is accumulated. A residual is reused while the accumulated value remains below 0.20; otherwise, a full transformer evaluation is performed and the accumulator is reset.

\section{General-80 Prompt Set}
\label{app:general80}
General-80 contains a General-40 split and an AV-hard-40 split. General-40 covers natural scenes, cities, people, animals, sports, cooking, industry, transportation, stage performances, and dynamic physical events. AV-hard-40 is balanced across the eight categories in Table~\ref{tab:app_general80}, with five prompts per category.

\begin{table}[htbp]
\centering
\caption{Composition of General-80.}
\label{tab:app_general80}
\small
\begin{tabular}{@{}lc@{}}
\toprule
\textbf{Split or AV-hard category} & \textbf{Prompts} \\
\midrule
General & 40 \\
Speech and lip synchronization & 5 \\
Music and pitch & 5 \\
Environmental sound & 5 \\
Physical-event synchronization & 5 \\
Animal vocalization & 5 \\
Mechanical sound and motion & 5 \\
Crowd ambience & 5 \\
Multiple audio-visual events & 5 \\
\midrule
Total & 80 \\
\bottomrule
\end{tabular}
\end{table}

Prompts were selected to provide broad visual coverage, observable sound sources or event times, balanced AV-hard categories, and a mixture of single-event and multi-event scenes. The list is deterministic and does not depend on online sampling. The complete prompts and split identifiers are provided in the anonymous code artifact at \path{scripts/ablation_prompts.py}; \texttt{general40}, \texttt{avhard40}, and \texttt{ablation80} reproduce the three corresponding selections.

\section{Evaluation Metrics and Evaluators}
\label{sec:appendix_metrics}

\subsection{VBench Visual Metrics}
We use the VBench implementation~\cite{huang2024vbench}. Table~\ref{tab:app_vbench_metrics} lists the six reported visual dimensions and their evaluators.

\begin{table}[htbp]
\centering
\caption{VBench visual metrics.}
\label{tab:app_vbench_metrics}
\small
\resizebox{\linewidth}{!}{
\begin{tabular}{@{}lll@{}}
\toprule
\textbf{Metric} & \textbf{Model} & \textbf{Definition} \\
\midrule
Subject consistency & DINO ViT-B/16 & Mean non-negative feature similarity to the first and preceding frames \\
Background consistency & CLIP ViT-B/32 & Analogous temporal similarity using CLIP image features \\
Motion smoothness & AMT-S & Agreement between interpolated and observed intermediate frames \\
Dynamic degree & RAFT & Fraction of videos exceeding the resolution-adjusted optical-flow criterion \\
Aesthetic quality & CLIP ViT-L/14 + LAION predictor & Mean predicted aesthetic score over sampled frames \\
Imaging quality & MUSIQ-SPAQ & No-reference per-frame quality, scaled by VBench to $[0,1]$ \\
\bottomrule
\end{tabular}
}
\end{table}

All six values are reported with a higher-is-better convention. Dynamic degree measures the presence of motion and should not, by itself, be interpreted as overall perceptual quality.

\subsection{Audio Quality and MLLM Evaluators}
\paragraph{DNSMOS.}
We use the DNSMOS P.835 local implementation with \texttt{sig\_bak\_ovr.onnx} and \texttt{model\_v8.onnx}. Audio is converted to mono 16\,kHz, and the reported value is the calibrated overall MOS estimate (higher is better).

\begingroup\sloppy
\paragraph{CLAP.}
We use LAION \path{clap-htsat-unfused}~\cite{elizalde2023clap}. Audio is converted to mono 48\,kHz and compared with the audio prompt. The score is\par\endgroup
\begin{equation}
 S_{\mathrm{CLAP}}=\cos(e_{\mathrm{text}},e_{\mathrm{audio}}).
\end{equation}

\paragraph{Qwen2.5-Omni evaluators.}
The Audio Reality, Visual Reality, and Expressiveness dimensions follow the VABench evaluator protocol~\cite{hua2025vabench} with Qwen2.5-Omni-7B. Decoding is deterministic (\path{do_sample=False}). Videos are sampled at 1 FPS with 4--8 frames and at most 150,528 input pixels. Each evaluator returns one integer score from 1 to 5 and a timestamped reason in a fixed JSON object. Audio Reality evaluates scale--pitch, material--timbre, and action-energy consistency; Visual Reality evaluates object permanence, biomechanics, rendering, temporal coherence, and material interaction; Expressiveness evaluates emotional alignment and narrative function. The exact system prompts and output schema used for evaluation are included in the anonymous code artifact at \path{VABench/vabench/third_audio_reality.py}, \path{VABench/vabench/third_visual_reality.py}, and \path{VABench/vabench/third_expressiveness.py}.

\subsection{Audio-Visual Alignment and Physics}
\paragraph{AV Sync.}
SyncFormer produces an absolute temporal offset $|\hat{o}_{\mathrm{SyncFormer}}|$ in seconds. The raw offset is lower-is-better. For the unified tables, we map it to
\begin{equation}
S_{\mathrm{AVSync}}
=
\max\left(0,1-\frac{|\hat{o}_{\mathrm{SyncFormer}}|}{0.5}\right).
\label{eq:app_avsync}
\end{equation}
The transformed score lies in $[0,1]$ and is higher-is-better; zero offset maps to 1, and offsets of 0.5 seconds or more map to 0.

\paragraph{VideoPhy2 / Lo-Phy.}
We use the VideoPhy2 evaluator distributed with AVGen-Bench~\cite{avgenbench}. It uniformly samples 32 frames, uses deterministic decoding, and returns a 1--5 assessment of local physical plausibility. The table reports the arithmetic mean (higher is better).

\paragraph{AV-F1.}
AV-F1 is a no-reference onset-alignment diagnostic. Visual onsets are local peaks in frame difference above the 75th percentile; audio onsets are local peaks in STFT spectral flux above the 75th percentile. Events are matched bidirectionally within $\pm3$ video frames. With the resulting precision $P$ and recall $R$,
\begin{equation}
 S_{\mathrm{AV\text{-}F1}}=\frac{2PR}{P+R},
\end{equation}
with a score of zero when $P+R=0$. Higher values indicate greater agreement between detected audio and visual event times; this handcrafted diagnostic is not a learned semantic evaluator.

\subsection{Pure-Pixel No-Reference Diagnostics}
The pure-pixel diagnostics resize the long side to at most 480 pixels and evaluate at most 128 frames. Except for the audio onset extraction used by AV-F1, they require no pretrained weights. Table~\ref{tab:app_pixel_metrics} summarizes their definitions.

\begin{table}[htbp]
\centering
\caption{Pure-pixel no-reference diagnostics.}
\label{tab:app_pixel_metrics}
\small
\resizebox{\linewidth}{!}{
\begin{tabular}{@{}lll@{}}
\toprule
\textbf{Metric} & \textbf{Definition} & \textbf{Direction} \\
\midrule
Sharpness & Variance of the grayscale Laplacian & Higher \\
Colorfulness & Hasler--Suesstrunk opponent-color statistic & Higher \\
Naturalness & Luminance/gradient kurtosis proximity to reference values & Higher \\
Temporal consistency & $1/(1+\operatorname{std}(d)/(\operatorname{mean}(d)+\varepsilon))$ for frame differences $d$ & Higher \\
Flicker & Mean second difference of frame-average brightness changes & Lower \\
Motion smoothness & Consecutive Farneback-flow directional agreement, mapped to $[0,1]$ & Higher \\
Artifact ratio & Fraction of $8\!\times\!8$ patches with local-variance $z$-score above 3 & Lower \\
AV-F1 & Bidirectional onset matching within $\pm3$ frames & Higher \\
\bottomrule
\end{tabular}
}
\end{table}

\section{Audio Cache Trigger Curves}
\label{app:trigger}
CAVC monitors the audio denoiser-output change $\Delta_a^t$ and activates caching after it remains below threshold $\theta$ for $W$ consecutive execution steps. Diffusion time $t$ decreases from $T$ to $0$, whereas the forward execution index $k$ used in the plots and trigger statistics increases. This is an online cacheability criterion, not a claim that audio numerically converges faster than video.

\begin{figure*}[t]
    \centering
    \includegraphics[width=0.95\textwidth]{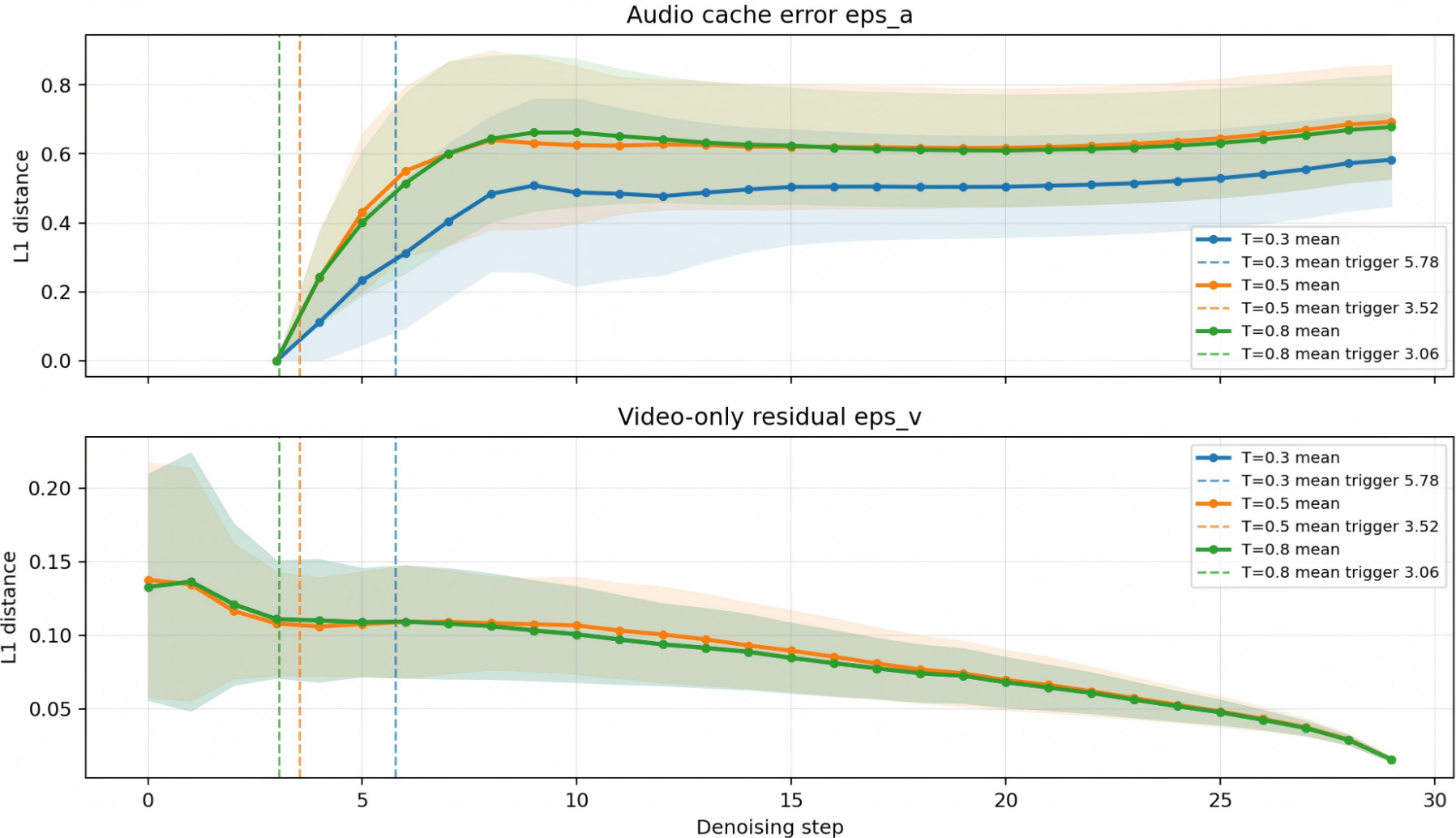}
    \caption{Empirical trigger curves over the forward execution index $k$ for threshold values 0.30, 0.50, and 0.80. Solid lines show the mean instantaneous audio cache residual (top) and video-only residual (bottom), shaded regions show dispersion, and vertical dashed lines mark the corresponding mean trigger indices $k_\tau$. The reported CAVC configuration uses $\theta=0.30$ and $W=3$.}
    \label{fig:app_trigger_curves}
\end{figure*}

\begin{table}[htbp]
\centering
\caption{Observed trigger statistics for the default configuration $\theta=0.30,W=3$. Trigger locations are reported using the forward execution index $k_\tau$.}
\label{tab:appendix_trigger}
\small
\resizebox{\linewidth}{!}{
\begin{tabular}{@{}lccccc@{}}
\toprule
\textbf{Benchmark} & \textbf{Median $k_\tau$} & \textbf{Mean $k_\tau$} & \textbf{Cached ratio} & \textbf{CAVC time (s)} & \textbf{Speedup} \\
\midrule
AVGen-Bench & 4 & 5.29 & 82.4\% & 329.69 & 1.10$\times$ \\
VABench T2AV & 4 & 4.91 & 83.6\% & 392.73 & 1.35$\times$ \\
\bottomrule
\end{tabular}
}
\end{table}

\section{Validity of Cross-Noise-Level Prediction Reuse}
\label{app:reuse}
In Phase B, the cached audio denoiser output $\hat{y}_a^*=\hat{y}_a^\tau$ is reused at lower noise levels $t<\tau$, while the scheduler continues to update the audio latent:
\begin{equation}
 \tilde a_{t-1}=S_t^a(\tilde a_t,\hat y_a^*).
\end{equation}
Thus, CAVC does not freeze the audio latent or the noise schedule. Let the full prediction residual during the cached interval satisfy $\|\hat y_a^t-\hat y_a^*\|\le\epsilon_a$, and assume that $S_t^a$ is locally Lipschitz in its latent and prediction arguments with constants $\alpha_t$ and $\beta_t$. The latent deviation then obeys
\begin{equation}
 e_a^{t-1}\le\alpha_t e_a^t+\beta_t\epsilon_a,
\end{equation}
and, because the full trigger-step update gives $e_a^{\tau-1}=0$,
\begin{equation}
 e_a^{t_{\mathrm{ref}}}\le\sum_{t=t_{\mathrm{ref}}+1}^{\tau-1}
 \left(\prod_{j=t_{\mathrm{ref}}+1}^{t-1}\alpha_j\right)\beta_t\epsilon_a.
\end{equation}
This conditional bound characterizes how prediction reuse affects the scheduler trajectory. It does not assert that the cached direction is exact; its practical suitability is evaluated by the residual curves and terminal-refinement ablations.

\section{Detailed Theoretical Derivation of the Output Deviation Bound}
\label{sec:appendix_derivation}

This section derives a conditional deviation bound for CAVC. The result depends on bounded prediction residuals and local Lipschitz assumptions for the scheduler, the video-only network, terminal joint updates, decoders, and the scalar evaluation function. It characterizes how approximation errors can propagate; it is not an unconditional perceptual-quality guarantee.

\subsection{Problem Formulation and Assumptions}
We formalize the exact joint denoising process and the approximated CAVC process to define the error metrics.

\noindent\textbf{Definition of Variables.}
Let $v_t$ and $a_t$ denote the exact video and audio latents at denoising step $t\in\{T,T-1,\ldots,1\}$, generated by the standard full joint workflow using the exact models $f_v(t) = f_v(v_t, a_t, c, \sigma_t)$ and $f_a(t) = f_a(v_t, a_t, c, \sigma_t)$. Denoising maps the state at $t$ to that at $t-1$ and terminates at $t=0$.
Conversely, let $\tilde{v}_t$ and $\tilde{a}_t$ denote the approximated latents produced by the CAVC framework. We quantify the cumulative deviation at step $t$ using the $L_2$ norms:
\begin{equation}
    e_a^t = \|a_t - \tilde{a}_t\|, \qquad e_v^t = \|v_t - \tilde{v}_t\|
\end{equation}

\noindent\textbf{Bounded-Residual Assumptions.}
Let $\mathcal{I}=\{\tau-1,\tau-2,\ldots,t_{\mathrm{ref}}+1\}$ denote the cached interval and let $\hat{y}_a^*$ be the full audio prediction cached at trigger step $\tau$. We assume that finite constants $\epsilon_a$ and $\epsilon_v$ exist over $\mathcal{I}$ such that
\begin{equation}
    \|f_a(t) - \hat{y}_a^*\| \le \epsilon_a
\end{equation}
Let $g_v(t) = g_v(v_t, c, \sigma_t)$ denote the isolated video network where audio cross-attention modules are explicitly bypassed. The bounded \textit{video-only residual} incurred by removing audio conditioning is defined as:
\begin{equation}
    \|f_v(v_t, a_t, c, \sigma_t) - g_v(v_t, c, \sigma_t)\| \le \epsilon_v
\end{equation}

\noindent\textbf{Lipschitz Constraints.}
We assume the updating scheduler $S_t(x, p)$ satisfies local Lipschitz continuity with respect to both the latent state $x$ (with constant $\alpha_t$) and the network prediction $p$ (with constant $\beta_t$):
\begin{align}
    \|S_t(x, p) - S_t(x', p)\| &\le \alpha_t \|x - x'\| \label{eq:lip_state} \\
    \|S_t(x, p) - S_t(x, p')\| &\le \beta_t \|p - p'\| \label{eq:lip_pred}
\end{align}

These inequalities are assumptions over the evaluated trajectory, not consequences of the trigger threshold alone. Figure~\ref{fig:app_trigger_curves} empirically measures the corresponding residual behavior.

\subsection{Single-Step Deviation Recurrence}
We derive the propagation of errors from step $t$ to $t-1$ for both modalities independently.

\noindent\textbf{Audio Branch Single-Step Deviation.}
Expanding the definition of the audio error at $t-1$:
\begin{equation}
    e_a^{t-1} = \|S_t^a(a_t, f_a(t)) - S_t^a(\tilde{a}_t, \hat{y}_a^*)\|
\end{equation}
To decouple the error caused by the drifted state $\tilde{a}_t$ from the error caused by the cached prediction $\hat{y}_a^*$, we introduce an intermediate virtual state $S_t^a(\tilde{a}_t, f_a(t))$ and apply the triangle inequality:
\begin{equation}
\begin{split}
    e_a^{t-1} &\le \|S_t^a(a_t, f_a(t)) - S_t^a(\tilde{a}_t, f_a(t))\| \\
    &\quad + \|S_t^a(\tilde{a}_t, f_a(t)) - S_t^a(\tilde{a}_t, \hat{y}_a^*)\|
\end{split}
\end{equation}
Applying the Lipschitz constraints (Eq. \ref{eq:lip_state} and \ref{eq:lip_pred}):
\begin{equation}
    e_a^{t-1} \le \alpha_t \|a_t - \tilde{a}_t\| + \beta_t \|f_a(t) - \hat{y}_a^*\|
\end{equation}
Substituting the bounds, we obtain the audio recurrence:
\begin{equation}
    e_a^{t-1} \le \alpha_t e_a^t + \beta_t \epsilon_a
    \label{eq:recurrence_audio}
\end{equation}

\noindent\textbf{Video Branch Single-Step Deviation (Nested Bounding).}
The derivation for the video branch is more intricate because the network structure itself changes from $f_v$ to $g_v$. The deviation is:
\begin{equation}
    e_v^{t-1} = \|S_t^v(v_t, f_v(t)) - S_t^v(\tilde{v}_t, g_v(\tilde{v}_t))\|
\end{equation}
Inserting the intermediate state $S_t^v(\tilde{v}_t, f_v(t))$ and applying the triangle inequality yields:
\begin{equation}
    e_v^{t-1} \le \alpha_t e_v^t + \beta_t \|f_v(t) - g_v(\tilde{v}_t)\|
\end{equation}
The critical challenge is evaluating the prediction discrepancy $\|f_v(t) - g_v(\tilde{v}_t)\|$, which encapsulates two distinct sources of error: the architectural change (removing audio) and the cascading effect of inputting a drifted state $\tilde{v}_t$ into the altered network $g_v$. 

We disentangle these by inserting another intermediate term, $g_v(v_t)$, applying the triangle inequality a second time:
\begin{equation}
    \|f_v(t) - g_v(\tilde{v}_t)\| \le \|f_v(t) - g_v(v_t)\| + \|g_v(v_t) - g_v(\tilde{v}_t)\|
\end{equation}
The first term is exactly the structurally induced video-only residual, bounded by $\epsilon_v$. For the second term, assuming the isolated video network $g_v$ is locally Lipschitz continuous with constant $L_g$:
\begin{equation}
    \|g_v(v_t) - g_v(\tilde{v}_t)\| \le L_g \|v_t - \tilde{v}_t\| = L_g e_v^t
\end{equation}
Consolidating these gives the total prediction discrepancy:
\begin{equation}
    \|f_v(t) - g_v(\tilde{v}_t)\| \le \epsilon_v + L_g e_v^t
\end{equation}
Substituting this back into the scheduler recurrence, we derive the final single-step bound for the video branch:
\begin{equation}
    e_v^{t-1} \le (\alpha_t + \beta_t L_g) e_v^t + \beta_t \epsilon_v
    \label{eq:recurrence_video}
\end{equation}
This formulation explicitly demonstrates that the video error amplification factor is increased to $(\alpha_t + \beta_t L_g)$, reflecting the compounding penalty of evaluating a drifted state through the approximated network.

\subsection{Multi-Step Accumulation}
We unroll the recurrences over $\mathcal{I}$. Because Phase A includes the full update at trigger step $\tau$, the exact and approximate states agree at the beginning of Phase B, so $e_a^{\tau-1}=e_v^{\tau-1}=0$. The audio deviation at the end of Phase B is
\begin{equation}
    e_a^{t_{\mathrm{ref}}} \le
    \sum_{t=t_{\mathrm{ref}}+1}^{\tau-1}
    \left( \prod_{j=t_{\mathrm{ref}}+1}^{t-1} \alpha_j \right)
    \beta_t \epsilon_a.
\end{equation}
Here, an empty product is defined as one.
Let $h_t$ denote the discrete integration step size and additionally assume $\alpha_t\le1+Lh_t$ and $\beta_t\le Bh_t$, with $H=\sum_{t=t_{\mathrm{ref}}+1}^{\tau-1}h_t$. Using $1+x\le\exp(x)$ gives

\begin{equation}
    e_a^{t_{\mathrm{ref}}} \le B \epsilon_a H \exp(LH),
    \label{eq:bound_audio}
\end{equation}

We apply the identical logic to the video recurrence (Eq. \ref{eq:recurrence_video}). The effective state amplification factor is bounded by:
\begin{equation}
    \alpha_t + \beta_t L_g \le 1 + (L + B L_g) h_t
\end{equation}
and therefore
\begin{equation}
    e_v^{t_{\mathrm{ref}}} \le B \epsilon_v H \exp((L + B L_g)H).
    \label{eq:bound_video}
\end{equation}

\subsection{Terminal Joint Propagation}
Phase C applies the same full joint update to the exact and approximate states. Define the product-state deviation $E_t=e_a^t+e_v^t$ and assume that each terminal joint update is locally Lipschitz in this norm with constant $\kappa_t$. For the terminal interval $\mathcal{J}=\{t_{\mathrm{ref}},t_{\mathrm{ref}}-1,\ldots,1\}$,
\begin{equation}
 E_0 \le K_C E_{t_{\mathrm{ref}}}, \qquad K_C=\prod_{t\in\mathcal{J}}\kappa_t.
 \label{eq:phase_c_bound}
\end{equation}
The bound allows either contraction or amplification; the empirical recoverability experiment in Table~\ref{tab:recoverability} measures the behavior of the actual terminal updates.

\subsection{Output Deviation Bound}
Let the video and audio decoders, $D_v$ and $D_a$, and a scalar evaluation function $Q$ be locally Lipschitz on the compared trajectories with constants $L_D^v,L_D^a$, and $L_Q$, respectively.

The final scalar-output discrepancy between the full sampler and CAVC is
\begin{equation}
\begin{split}
    |Q(\text{full}) - Q(\text{CAVC})| &= \\
    |Q(D_v(v_0), &D_a(a_0)) - Q(D_v(\tilde{v}_0), D_a(\tilde{a}_0))|
\end{split}
\end{equation}
Applying the triangle inequality and the stated local assumptions yields
\begin{equation}
    |Q(\text{full}) - Q(\text{CAVC})| \le L_Q \Big( L_D^v e_v^0 + L_D^a e_a^0 \Big)
    \label{eq:final_bound}
\end{equation}

Combining Equations~\ref{eq:bound_audio}--\ref{eq:phase_c_bound} with Equation~\ref{eq:final_bound} makes the dependence on $\epsilon_a$, $\epsilon_v$, cached duration $H$, and the assumed local constants explicit. This is an output-deviation characterization under the listed assumptions. It neither guarantees an improvement in any perceptual metric nor rules out mode collapse or other failure modes.

\end{document}